\documentclass{svmult}

\usepackage{makeidx}         
\usepackage{graphicx}        
\usepackage{multicol}        
\usepackage[bottom]{footmisc}
\usepackage[T1]{fontenc}
\usepackage[cp1250]{inputenc}

\makeindex             

\begin{document}

\title*{Vehicles recognition using fuzzy descriptors \newline of image segments}
\titlerunning{Vehicles recognition using fuzzy descriptors of image segments}
\author{Bart{\l}omiej P{\l}aczek}
\institute{Faculty of Transport, 
Silesian University of Technology \newline
ul. Krasinskiego 8, 40-019 Katowice, Poland \newline
\texttt{bartlomiej.placzek@polsl.pl}}
%
%
\maketitle
\begin{abstract}
In this paper a vision-based vehicles recognition method is presented. Proposed method uses  fuzzy description of image segments for automatic  recognition of vehicles recorded in image data. The description takes into account selected geometrical properties and shape coefficients determined for segments of reference image (vehicle model).
The proposed method was implemented using reasoning system with fuzzy rules. A vehicles recognition algorithm was developed based on the fuzzy rules describing shape and arrangement of the image segments that correspond to visible parts of a vehicle. An extension of the algorithm with set of fuzzy rules defined for different reference images (and various vehicle shapes) enables vehicles classification in traffic scenes. The devised method is suitable for application in video sensors for road traffic control and surveillance systems.

Preprint of: P{\l}aczek, B.,  Vehicles Recognition Using Fuzzy Descriptors of Image Segments. In: Kurzynski M. et al. (eds.) Computer Recognition Systems 3, Advances in Intelligent and Soft Computing , vol. 57/2009, pp. 79--86. Springer-Verlag, Berlin Heidelberg (2009). 
The final publication is available at www.springerlink.com
\end{abstract}

\section{Introduction}
\label{idauthor:sec:1}

Vehicles recognition is an important task for vision-based sensors. Accomplishment of this task enables advanced traffic control algorithms to be applied. E.g. recognition of public transport vehicles (buses, trams) allows for priority introducing in traffic signals control. Recognition of personal cars, vans, trucks etc. enables determination of optimal assignment of green time for particular crossroad approach. It is obvious that traffic control algorithms using additional information on vehicles classes provides lower delays and higher fluidity level of the traffic in comparison to less sophisticated control strategies.

For real-world traffic scenes vehicles recognition is a complex problem. Difficulties are encountered especially for crossroads, where queues are formed and vehicles under scrutiny are occluded by dynamic or stationary scene components. Vehicles recognition systems have also to cope with influence of objects shadows and ambient light changes.

In this paper a method is introduced that allows for vehicles recognition on the basis of simple image segments analysis. Matching of segments detected in the image with model segments is performed using fuzzy reasoning system. For the proposed method particular image segments are analysed that correspond to visible parts of a vehicle. This approach enables recognition of partially occluded vehicles. Applied method of segments description includes several shape parameters and enables considerable reduction of data amount processed by the reasoning system.

Rest of this paper is organized as follows. Section 2 includes a brief survey of vision-based vehicles recognition methods and presents a general design of the proposed method. Section 3 includes definitions of segments parameters and merging operation. Section 4 describes fuzzy rules designed for vehicles recognition. In section 5 implementation details of the fuzzy reasoning system are discussed. Section 6 deals with pre-processing of the input data, including segments extraction and merging as well as application of vehicle models. An experiment results are shown in section 7 and finally, conclusions are drawn.

\section{Related works and proposed method}
\label{idauthor:sec:2}

Several vision-based vehicles recognition methods have been developed so far. All of these methods use various forms of vehicle shape models. Some of the models describe shape of entire vehicle body (3-D models); other models consider selected shape properties or local features arrangement in the vehicle image. Usually operations of vehicle detection and tracking are preliminary steps in the algorithm of vehicles recognition.

In \cite{idauthor:journal3} a method is introduced that uses only vehicle dimensions to classify vehicles into two categories: cars and non-cars (vans, pickup, trucks, buses, etc.). Vehicles models are defined as rectangular image regions with certain dynamic behavior. Correspondences between regions and vehicles are recognized on the basis of tracking results.

A vehicle recognition approach that uses parameterised 3-D models is described in \cite{idauthor:journal2}. The system uses a generic polyhedral model based on typical shape to classify vehicles in a traffic video sequence. Using 3-D models partially occluded vehicles can be correctly detected \cite{idauthor:journal4}. However, algorithms of this type have higher computational complexity.

Feature-based vehicle recognition algorithms \cite{idauthor:journal5} include edge detection techniques, corner detection, texture analysis, harr-like features, etc. On the basis of these detection tools the vehicle  recognition methods have been developed \cite{idauthor:journal9}. In \cite{idauthor:journal6} a vehicles classifier is proposed based on local-feature configuration. It was applied to distinguish four classes: sedan, wagon, mini-van and hatchback.

Another approach combines elements of 3-D models with feature-based methods. Local 3D curve-groups (probes) are used, which when projected into video frames are features for recognizing vehicle in video sequence \cite{idauthor:journal1}. This classifier was applied to three classes of vehicles: sedans, minivans and SUV's.

The method introduced in this paper was designed to recognise vehicles and categorize them into classes that are relevant from a point of view of  the traffic control objectives. Taking into account impact on a crossroad capacity and priority level, five fundamental vehicle classes have to be distinguished: personal car, van, truck, bus and tractor-trailer.

In the proposed approach vehicles recognition is performed by reasoning procedure using fuzzy rules \cite{idauthor:journal8}. The rules describe shape of a vehicle model and allow for level evaluation of similarity between image objects and the assumed model. Scheme of the method is presented in fig. 1. The recognition is based on description matching of the model segments with segments extracted from an input image. Segment description takes into account its geometrical properties, shape coefficients and location. This particular data format was applied for both model segments (reference image) as well as segments extracted from input image. Processing procedure of input image includes segmentation and merging of the extracted segments (see sec. 6).
This method do not require tracking data to recognise a vehicle. The complexity reduction of recognition algorithm is achieved due to processing of simplified image description. 

\begin{figure}
\centering
\includegraphics[height=3.2cm]{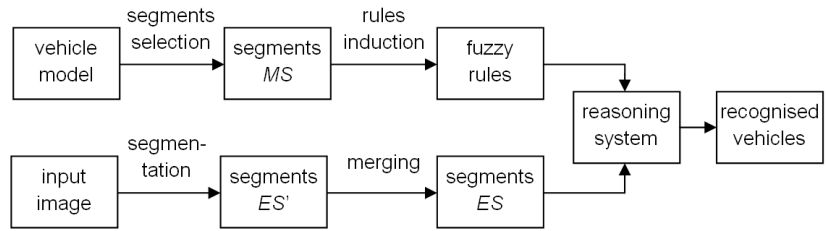}
\caption{Scheme of the vehicles recognition method}
\label{idauthor:fig:1}      
\end{figure}

Fuzzy rules are inducted on the basis of segments properties determined for vehicle model. Rules induction process consists in fuzzyfication of the parameters describing model segments $M\!S$. Antecedents of the rules take into account description of segments $M\!S$; consequents indicate degree of similarity between input image and the model.
 
The segments merging, rules induction and reasoning procedures are performed using devised representation method of image segments (segments description based on selected geometric parameters). Operations are not executed directly on image data, what is crucial in reducing computational complexity of the recognition algorithm. These procedures will be described in detail in the following sections.

\section{Segments description and merging operation}
\label{idauthor:sec:3}

Introduced shape description method for objects registered in an image is based on selected geometrical parameters of the image segments. The description method applied for segments of reference image (model) as well as for extracted segments of the input image remains the same.

Description of $i$-th image segment is defined by the formula:
\begin{equation}
S_{i}=\left(A_{i}, x_{i}, y_{i}, x^{min}_{i}, x^{max}_{i}, y^{min}_{i}, y^{max}_{i}\right),
\end{equation}
where: $A_{i}$ - area of the segment, $c_{i}=\left(x_{i},y_{i}\right)$ - centre of mass determined for the segment, $w(S_{i})=x^{max}_{i}-x^{min}_{i}$ - width of the segment, $h(S_{i})=y^{max}_{i}-y^{min}_{i}$ - height of the segment.

Computations of the segments parameters, defined above, are performed for coordinate system x-y (fig. 2) that has orientation determined by the model orientation in camera coordinate system.

\begin{figure}
\centering
\includegraphics[height=4.8cm]{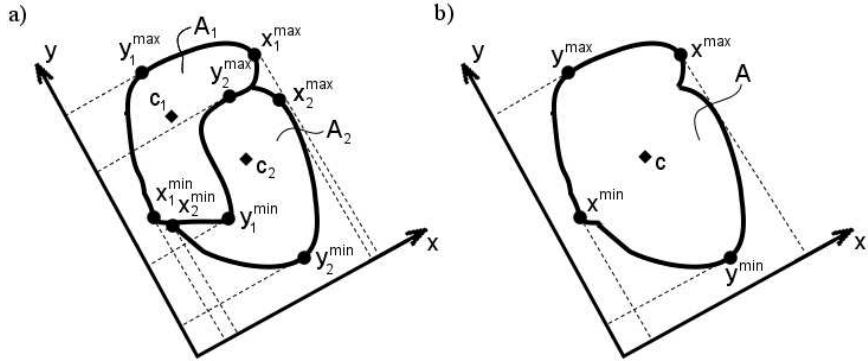}
\caption{Segments merging operation}
\label{idauthor:fig:2}      
\end{figure}

The principles of segments merging operation are illustrated in fig. 2. Parameters of merged segments are depicted in fig. 2 a); fig. 2b) presents results of the segments merging. 
Formally, the merging operation is denoted by relation:
\begin{equation}
S_{k}=S_{i}\hspace{2pt}^{M} S_{j},
\end{equation}
which means that segment $S_{k}$ is created by merging $S_{i}$ with $S_{j}$ and parameters of its description are computed according to the following formulas:
\begin{eqnarray}
A_{k}=A_{i}+A_{j},\hspace{10pt}x_{k}=\frac{A_{i} x_{i}+A_{j} x_{j}}{A_{i}+A_{j}},\hspace{10pt}y_{k}=\frac{A_{i} y_{i}+A_{j} y_{j}}{A_{i}+A_{j}},\nonumber\\
x^{min}_{k}=\min\left(x^{min}_{i},x^{min}_{j}\right),\hspace{10pt} y^{min}_{k}=\min\left(y^{min}_{i}, y^{min}_{j}\right),\\
x^{max}_{k}=\max\left(x^{max}_{i},x^{max}_{j}\right),\hspace{10pt} y^{max}_{k}=\max\left(y^{max}_{i}, y^{max}_{j}\right).\nonumber
\end{eqnarray}
The merging operation has following properties that are important for its implementation:
\begin{equation}
S_{i}\hspace{2pt}^{M} S_{j}=S_{j}\hspace{2pt}^{M} S_{i},\hspace{10pt}
S_{i}\hspace{2pt}^{M} \left(S_{j}\hspace{2pt}^{M} S_{k}\right)=\left(S_{i}\hspace{2pt}^{M} S_{j}\right)\hspace{2pt}^{M} S_{k}.
\end{equation}

Merging operation is used for pre-processing of segments extracted in input image. It is necessary for correct vehicle recognizing by fuzzy rules that are introduced in the following section.

\section{Fuzzy rules for vehicles recognition}
\label{idauthor:sec:4}

Fuzzy rules are produced for recognition procedure on the basis of segments defined in reference image (model). As the method was intended to categorize vehicles into classes, a separate model and rules are needed for each class.

Set $M\!S^{c}$ includes descriptions of all segments determined for vehicle model of class $c$ : $M\!S^{c}=\left\{S^{c}_{j}\right\},j=1...n \left(c\right)$
Segments extracted from input image are described by elements of set $E\!S$ :
$E\!S=\left\{S_{i}\right\},i=1...m$.

For every vehicles class a set of classification rules is created including three types of rules, connected with different aspects of the segments layout: shape rules, placement rules and arrangement rules.

Shape rules (5) describe shape of segments defined for the vehicle model of class $c$ using shape coefficients. They allow for similarity determination of segments from $M\!S^{c}$ and $E\!S$.
\begin{equation}
\textbf{if } A_{i} \mbox{ is } \tilde{A^{c}_{j}} \textbf{ and } q(S_{i}) \mbox{ is } \tilde{q}(S^{c}_{j})
\textbf{ then } p(S_{i}) \mbox{ is } j,
\end{equation}
where: $i=1...m$, $j=1...n(c)$ and $q(S_{i})=w(S_{i})/h(S_{i})$ is shape coefficient.  

Placement rules (6) describe mutual placement of the segments. Using these rules each pair of segments from $E\!S$ is processed. It is performed by comparing the relative location of their mass centres (defined by $[dx, dy]$ vector) with the mass centres arrangement of the model segments $(j_{1}, j_{2})$.
\begin{eqnarray}
\textbf{if } dx(S_{i_1},S_{i_2}) \mbox{ is } \tilde{dx}(S^c_{j_1},S^c_{j_2}) 
\textbf{ and } dy(S_{i_1},S_{i_2}) \mbox{ is } \tilde{dy}(S^c_{j_1},S^c_{j_2})\nonumber\\
\textbf{ and } p(S_{i_2}) \mbox{ is } j_2
\textbf{ then } d(S_{i_1},S^c_{j_2}) \mbox{ is } (j_1,j_2),
\end{eqnarray}
where: $i_1, i_2=1...m$, $i_1 \neq i_2$, $j_1=1$, $j_2=2...n(c)$, $dx(S_i,S_j)=x_i-x_j$, $dy(S_i,S_j)=y_i-y_j.$

The spatial arrangement rules (7) take into account placement of all model segments. Both the segments shape similarity and segments locations are checked. It is performed using results of the previous reasoning stages that exploit (5) and (6) rules.
\begin{eqnarray}
\textbf{if } p(S_i) \mbox{ is } 1 \textbf{ and } d(S_i,S^c_2) \mbox{ is } (1,2) 
\textbf{ and } d(S_i,S^c_3) \mbox{ is } (1,3)\nonumber\\
\textbf{ and } ... \textbf{ and } d(S_i,S^c_{n(c)}) \mbox{ is } (1,n(c))
\textbf{ then } class \mbox{ is } c,
\end{eqnarray}
where: $i=1...m$.

The tilde symbol was used in rules definitions (5) and (6) to indicate fuzzy sets having trapezoidal membership functions. The membership functions are determined for specific values of model segments parameters.
 
\section{Reasoning system implementation}
\label{idauthor:sec:5}
A registered vehicle is recognised using fuzzy reasoning system with base of rules that are defined in the previous section. Input data of the reasoning system is fuzzified into type-0 fuzzy sets (singleton fuzzification). The system uses Mamdani reasoning method and averaging operator (arythmetic mean) for fuzy sets aggregation \cite{idauthor:monograph1}.
Due to these assumptions, the membership functions for consequents of rules (5) - (7) are computed according to the formulas (8) - (10).

In the first stage of the reasoning procedure a membership function is computed for each segment $S_i$ that determines its similarity to particular segments $S^c_j$ of the model.
\begin{equation}
\mu_{p(S_i)}(j)=\frac{1}{2}\left[\mu_{\tilde{A}^c_j}(A_i)+\mu_{\tilde{q}(S^c_j)}(q(S_i))\right],
\hspace{5pt}i=1...m, j=1...n(c).
\end{equation}

In the second stage another membership function is evaluated to check if the segments arrangement in the image is consistent with the model definition:
\begin{eqnarray}
\mu_{d(S_i,S^c_{j_2})}(j_1,j_2)=\max_{i_2}\left\{\frac{1}{3}\left[\mu_{\tilde{dx}(S^c_{j_1},S^c_{j_2})}(dx(S_{i_1},S_{i_2}))+
\right.\right.\nonumber\\
+\left.\left.\mu_{\tilde{dy}(S^c_{j_1},S^c_{j_2})}(dy(S_{i_1},S_{i_2}))+\mu_{p(S_{i_2})}(j_2)\right]\right\},
\end{eqnarray}
where: $i_1=1...m$, $i_2=1...m$, $i_1 \neq i_2$, $j_1=1$, $j_2=2...n(c)$.

At last, the third membership function determines similarity level of the object recorded in input image and vehicle shape model of class $c$: 
\begin{equation}
\mu_{class}(c)=\max_i\left\{\frac{1}{n(c)+1}\left[\mu_{p(S_i)}(1)+\sum_j\mu_{d(S_i,S^c_j)}(1,j)\right]\right\},
\end{equation} 
where $i=1...m$.

The class number of recognised vehicle is a product of the defuzzification of function (10). It is determined by the maximal membership method.

\section{Processing of input data}
\label{idauthor:sec:6}
Input data of the vehicle recognition algorithm (fig. 1) consists of descriptors computed for both segments $E\!S$ detected in the input image as well as segments $M\!S$, selected on the basis of the vehicle model.

Segments $E\!S'$ in the input image can be extracted using various segmentation methods. In the presented approach background subtraction, edge detection and area filling algorithms were applied for this task \cite{idauthor:contribution1}. Thus, segments $E\!S'$ correspond to regions of input image that are bounded by edges and do not belong to the background.

Processing procedure of input image includes merging of the extracted segments $E\!S'$ into segments $E\!S$, that match better with shape of the model $M\!S$. The segments merging is motivated by assumed fidelity level of the model, which does not take into account minor parts of vehicle like headlights or number plate.
Set $E\!S'$ of extracted segments is transformed into $E\!S$. This operation uses threshold value $\tau$ of similarity between merged segments $S_i$ and model segments $S_j$:
\begin{equation}
E\!S = \left\{ S_i=\textbf{M}(P\!S_i):\exists \ S_j \in M\!S: \mu_{p(S_i)}(j)\geq\tau  \right\},
\end{equation}
$P\!S_i=\left\{S_{ik}\right\}$, $k=1...z$ is a set of extracted segments $P\!S_i \subset E\!S$ and $\textbf{M}(P\!S_i)$ denotes the merging operation on set $P\!S_i$: $\textbf{M}(P\!S_i)=S_{i1}\hspace{2pt}^{M}S_{i2}\hspace{2pt}^{M}...^{M} S_{iz}$.

The rules induction procedure requires parameters determination of the reference image (model) segments. A reference image description (set $M\!S$) is generated on the basis of three-dimensional (3-D) vehicle shape model that describes faces arrangement of a vehicle body \cite{idauthor:contribution1}. This model is transformed into 2-D model, for defined locations of the vehicle and camera. Thus, the 2-D model establishes a reference vehicle image of a given class that is located in a specific place within the scene. Faces of the 3-D model, projected on image plane, defines segments $M\!S$ of the model. The vehicle model has to be defined for each vehicles class that is taken into consideration by the recognition procedure.

\section{Experimental results}
\label{idauthor:sec:7}
The proposed vehicles recognition method was tested using number of images selected in video sequences of traffic scenes, where vehicles of different classes have been recorded. Furthermore, synthetic traffic images were utilised for vehicles queue modeling in performance analysis of the method.

The experiments of vehicle class recognition were performed using vehicle models for five classes (personal car, van, truck, bus and tractor-trailer). Segments determined in vehicles models were matched with those extracted from input images. This task was executed by applying fuzzy reasoning system described in sec. 5. 
Examples of the experiment results are presented in fig. 3. It includes input images and vehicle models marked with white lines. The models are displayed corresponding with recognised class of a vehicle, identified by maximal value of function $\mu_{class}$ (eq. 10). For particular cases depicted if fig. 3 the recognition results were obtained as follows: a) personal car 0,44; b) personal car 0,51; c) personal car 0,38; d) bus 0,37; e) truck 0,31; personal car 0,41; f) tractor-trailer 0,35. The above numbers denote maximum of $\mu_{class}$ for each example.
\begin{figure}
\centering
\includegraphics[height=2.6cm]{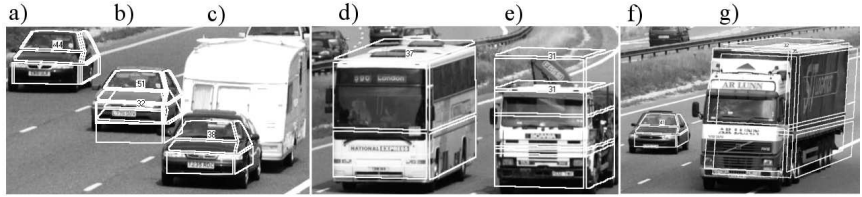}
\caption{Examples of recognised vehicles}
\label{idauthor:fig:3}      
\end{figure}

The overall result of the experiments indicates that for nearly 90\% of test images the proposed vehicle recognition method provides correct classification results. However, the results strongly depend on correctness of the segments extraction in input images. 

\section{Conclusion}
\label{idauthor:sec:8}
A fuzzy description method of image segments  was introduced. The method was implemented for automatic recognition of vehicles recorded in image data. This implementation was based on fuzzy reasoning system with rules describing properties of image segments.

Experimental results confirmed that the proposed method is effective to vehicle recognition. It was demonstrated that the system can categorize vehicles into five classes. It should be noticed that low complexity of the proposed image description makes the method suitable for application in video sensors for road traffic control and surveillance systems.

%
%

%
%



\end{document}